\RequirePackage{fix-cm}
\documentclass[smallcondensed]{svjour3}     
\smartqed  
\usepackage{graphicx}
\journalname{Machine Learning}
\usepackage{natbib}
\usepackage{pgfplots}
\pgfplotsset{compat=newest}
\pgfplotsset{plot coordinates/math parser=false}
\newlength\figureheight
\newlength\figurewidth
\usepackage{standalone}
\pgfplotsset{every axis legend/.append style={legend cell align=left}}

\usepackage{subfiles}
\makeatletter
\def\cl@chapter{\@elt {theorem}}
\makeatother
\usepackage{tikz}
\usepackage{subfig}
\usepackage{booktabs}
\usepackage{algorithm}
\usepackage[noend]{algpseudocode}

\usepackage{centernot}
\usetikzlibrary{arrows}
\usepackage{url}
\usepackage{paralist} 

\usepackage{pgfplots}

\pgfplotsset{compat=newest}
\pgfplotsset{every axis legend/.append style={%
cells={anchor=west}}
}
\usetikzlibrary{arrows}
\tikzset{>=stealth'}

\definecolor{color1}{RGB}{0, 0, 0}
\definecolor{color2}{RGB}{230, 159, 0}
\definecolor{color3}{RGB}{86, 180, 233}
\definecolor{color4}{RGB}{213, 94, 0}

\newcommand{\Chris}[1]{\marginpar{\textcolor{cyan}{Chris: #1}}}

\newcommand{\xhdr}[1]{{\noindent\bfseries #1}.}

\newcommand{\dom}[1]{\text{dom} #1}
\newcommand{\compl}[1]{#1^\complement}
\newcommand{\violated}{\textsc{Violated}\xspace}
\newcommand{\holds}{\textsc{Holds}\xspace}
\newcommand{\unknown}{\textsc{Unknown}\xspace}
\newcommand{\worsethanopt}{\textsc{WorseThanOpt}\xspace}
\newcommand{\optimal}{\textsc{Optimal}\xspace}

\usepackage{amsmath}
\usepackage{hyperref}
\usepackage{xcolor}
\colorlet{linkequation}{blue}
\usepackage[noabbrev]{cleveref}

\usepackage{booktabs}

\usepackage{siunitx}

\usepackage{amssymb}
\usepackage{amsfonts}       
\newcommand{\norm}[1]{\left\lVert#1\right\rVert}

\usepackage{booktabs}
\usepackage{subfig}
\usepackage{multicol}
\usepackage{tabularx}
\usepackage{multirow}
\usepackage{xspace}

\usepackage{vector} 
\renewcommand{\vec}[1]{\vect{#1}}
\newcommand{\mat}[1]{\vec{#1}}

\newcommand{\defn}[1]{\emph{#1}}
\newcommand{\searchandopt}{search with optimization\xspace}

\newcommand{\searchandreach}{search with reachability\xspace}
\newcommand{\Searchandreach}{Search with reachability\xspace}
\newcommand{\optimizer}{MarabouOpt\xspace}

\begin{document}


\title{ Global Optimization of Objective Functions Represented by ReLU Networks 
}

\author{Christopher A. Strong \and
        Haoze Wu \and
        Aleksandar Zeljić \and
        Kyle D. Julian \and
        Guy Katz \and
        Clark Barrett \and
        Mykel J. Kochenderfer}


\institute{C.A. Strong (corresponding author) \at
              Stanford University, Department of Electrical Engineering \\
              Stanford, CA, USA \\
              \email{christopher\_strong@berkeley.edu}            \\
              ORCID: 0000-0002-8914-6852
           \and
           H. Wu, A. Zeljić, C. Barrett \at
              Stanford University, Department of Computer Science \\
              Stanford, CA, USA \\
              \email{\{haozewu, zeljic, barrett\}@stanford.edu}           \\
              ORCID: 0000-0002-5077-144X, 0000-0003-0673-9327, 0000-0002-9522-3084
           \and 
           K.D. Julian, M.J. Kochenderfer \at
             Stanford University, Department of Aeronautics and Astronautics \\
             Stanford, CA, USA \\
             \email{\{kjulian3, mykel\}@stanford.edu} \\
             ORCID: 0000-0002-6247-1874, 0000-0002-7238-9663
           \and 
            G. Katz \at 
             The Hebrew University of Jerusalem, School of Computer Science and Engineering \\
             Jerusalem, Israel \\
             \email{guykatz@cs.huji.ac.il} \\
             ORCID: 0000-0001-5292-801X
}

\date{Received: date / Accepted: date}

\maketitle

\begin{abstract}

Neural networks can learn complex, non-convex functions, and it is challenging to guarantee their correct behavior in safety-critical contexts. Many approaches exist to find failures in networks (e.g., adversarial examples), but these cannot guarantee the absence of failures. Verification algorithms address this need and provide formal guarantees about a neural network by answering ``yes or no'' questions. For example, they can answer whether a violation exists within certain bounds. However, individual ``yes or no" questions cannot answer qualitative questions such as ``what is the largest error within these bounds''; the answers to these lie in the domain of optimization. Therefore, we propose strategies to extend existing verifiers to perform optimization and find: (i) the most extreme failure in a given input region and (ii) the minimum input perturbation required to cause a failure. A naive approach using a bisection search with an off-the-shelf verifier results in many expensive and overlapping calls to the verifier. Instead, we propose an approach that tightly integrates the optimization process into the verification procedure, achieving better runtime performance than the naive approach. We evaluate our approach implemented as an extension of Marabou, a state-of-the-art neural network verifier, and compare its performance with the bisection approach and MIPVerify, an optimization-based verifier. We observe complementary performance between our extension of Marabou and MIPVerify.

\end{abstract}

\section{Introduction}
\label{sec:intro}


Artificial deep neural networks (DNNs) have demonstrated great promise in a
wide variety of
applications~\citep{schmidhuber2015deep, liu2017survey}. These applications include
image recognition~\citep{krizhevsky2012imagenet}, control
\citep{hunt1992neural}, and natural language processing
\citep{otter2020survey}, among many others.  Because of these successes, there is naturally interest in incorporating DNNs into other applications, including safety-critical
systems~\citep{BoDeDwFiFlGoJaMoMuZhZhZhZi16,JuLoBrOwKo16}.
Although DNNs are obtaining unprecedented results, their opacity poses
significant challenges --- especially in the context of
safety-critical systems, where mistakes can endanger lives and cause
significant damage. A notable example includes DNNs in autonomous
driving systems, where unexpected behavior of the DNN could harm passengers or pedestrians. Consequently, it is especially
desirable to formally reason about DNNs, providing rigorous guarantees
about their behaviors.

Recent research has focused on \emph{neural network
  verification}~\citep{HuKwWaWu17, katz2017reluplex,GeMiDrTsChVe18,wang2018formal}. Verification
involves answering
``yes or no'' questions about DNNs, and can be used to rule out undesirable
behaviors. For example, a verification query for an autonomous driving DNN could ask whether an input exists that encodes a situation in which the autonomous vehicle is approaching an obstacle, but for which the DNN advises the vehicle to maintain the current course. If the verification engine
answers no, we are guaranteed that this particular behavior can never
happen for \emph{any possible input}. If it answers yes, then it returns an input that leads to the undesirable outcome. The verification problem has been shown to be NP-complete
\citep{katz2017reluplex}; however, large strides have been made
in recent years in solving networks that arise in practice~\citep{wang2018efficient, weng2018towards, katz2019marabou,tran2020nnv, wu2020parallelization}.

Although tremendous effort has been put into answering yes or no questions about DNNs, formally
answering quantitative questions about them has received less
attention. Such questions can be highly important when verifying a system: for example, we may want to know \emph{how close} an obstacle can
be before the DNN controller turns the vehicle, or \emph{how much} the steering of the car can be affected by small errors in the input image (e.g., caused by a malfunctioning camera). Finding answers to these questions requires an \emph{optimization} process. Existing techniques can provide bounds on the answers to these questions, but the bounds may be too loose to effectively reason about the performance of the system \citep{singh2018fast,wang2018efficient,wang2018formal,weng2018towards,zhang2018efficient,boopathy2019cnn,liu2019algorithms}. For example, knowing that the most extreme steering angle for a self-driving car is somewhere between \num{-40} and \num{80} degrees is likely not sufficiently informative to deduce guarantees about the car's behavior.

Global optimization of neural networks is also useful in the context of interpreting the patterns that a network has learned. For example, the activation of a hidden node can be maximized or minimized with respect to the input, in order to acquire a sense of the input properties that that node is tuned for. This has been done with local optimizers, but to our knowledge has not been explored with global optimizers \citep{le2013building, ribeiro2016should}.

Many approximate techniques for answering these questions use heuristics and local optimization to find inputs that lead to undesired behavior from the network.  Such techniques are referred to as \defn{adversarial attacks}, and the corresponding inputs that lead to the undesired behavior are known as \defn{adversarial examples}. Various techniques have been proposed, both for performing such attacks and defending against them~\citep{goodfellow2014explaining,carlini2017towards,
chakraborty2018adversarial,
yuan2019adversarial}. Additionally, a variety of techniques have been proposed to find certified lower bounds on the perturbation required to produce an adversarial example \citep{weng2018towards,zhang2018efficient,boopathy2019cnn}. Although approximate techniques typically scale much more effectively to large networks \citep{muller2020neural},
we may need a stronger guarantee on the behavior of a safety-critical system than we can find with existing approximate techniques.
Existing strategies to solve global optimization problems on neural
networks include performing a binary search for the optimal
value by repeatedly calling neural verifiers, as well as mixed integer
programming (MIP) approaches that explicitly encode the network
as constraints in an optimization problem. Repeatedly calling a neural
verifier is often computationally prohibitive, while MIP
approaches work well on some problem types but struggle with others \citep{carlini2017provably, tjeng2017evaluating}.
It is thus desirable to develop new approaches for better accommodating different problem domains.

In order to have a wider variety of approaches for neural optimization, we introduce a framework for converting existing verifiers into optimizers. We demonstrate our framework by extending the Marabou verifier, which implements the Reluplex algorithm \citep{katz2019marabou}. Our extended version of Marabou, which we refer to as \optimizer, performs a branch-and-bound search over the activation space of the network. We compare the runtime of our solver to MIPVerify, a MIP approach that also solves global optimization problems, and we find that the two approaches are complementary to each other on the benchmarks tested. Additionally, we compare the approximate optima found by adversarial attack algorithms to the true optima found by \optimizer.

This paper is organized as follows: \cref{sec:background} provides background and descriptions of the notation used in the rest of the paper; \cref{sec:convert verifiers} describes high-level approaches for modifying four categories of verifiers to perform optimization; \cref{sec:modifying marabou} describes in detail how the Reluplex algorithm can be modified to perform optimization; \cref{sec:results} presents our experimental setup and results; \cref{sec:related work} summarizes related work; and \cref{sec:conclusion} concludes and suggests future research directions. 


\section{Background and Problem Formulation}
\label{sec:background}

This section introduces notation and defines the neural network verification and neural network optimization problems. It also provides a categorization of neural network verification algorithms and explains existing local optimizers.
\\\\
\xhdr{Neural Networks}
We denote the function represented by a neural network $N$ with $n$ inputs and $m$ outputs as $f(\vec{x}): \mathbb{R}^n \to \mathbb{R}^m$. Let a network $N$ with $K$ layers, including the input and output layers, have input to layer $\ell$ denoted by $\hat{\vec{z}}_\ell$ and output of layer $\ell$ denoted by $\vec{z}_\ell$. The input to each layer (besides the first) is computed by applying an affine transformation to the previous layer followed by an \emph{activation function}.  We consider two activation functions in this paper: the identity function and the rectified linear unit (ReLU).  For a rectified linear unit (ReLU) layer $\ell$, we have
\begin{equation}
    \vec{z}_{\ell} = \max(0, \hat{\vec{z}}_\ell).
\end{equation}
For identity layers, we have 
\begin{equation}
    \vec{z}_{\ell} = \hat{\vec{z}}_\ell.
\end{equation}
Let $\mat{W}_\ell$ and $\vec{b}_\ell$ be the weights and biases connecting layer $\ell$ to layer $\ell + 1$ such that
\begin{equation}
    \hat{\vec{z}}_{\ell + 1} = \mat{W}_\ell\vec{z}_\ell + \vec{b}_\ell.
\end{equation}
The output of the network will be referred to as $\vec{y}$, with
\begin{equation}
    \vec{y} = \vec{z}_K.
\end{equation}

Let $\mathcal{L}$ denote the set of indices of ReLU layers and $\mathcal{I}$ denote the set of indices of identity layers. A ReLU is considered \defn{active} if its input is greater than or equal to 0, and \defn{inactive} otherwise. An \defn{activation state} is a representation of whether a node is active or not for each ReLU in a network. For a given activation state, let $\mathcal{A}$ be the set of indices $(i, j)$ of active ReLUs where $i$ is the layer and $j$ is the node in the layer. Similarly, let $\mathcal{N}$ be the set of indices $(i, j)$ of inactive ReLUs. A \defn{partial activation state} is an activation state where some nodes are left unknown. In this case, let $\mathcal{U}$ be the set of indices $(i, j)$ of undetermined nodes for the partial activation state.
\\\\
\xhdr{Geometric Objects}
 We refer to sets described by the intersection of affine inequalities as \defn{polytopes}. A polytope $\mathcal{P}$ can be described by a matrix $\mat{A}$ and vector $\vec{b}$ as $\mathcal{P} = \{\vec{x} : \mat{A}\vec{x} \le \vec{b}\}$. We refer to the complement of a polytope as a \defn{polytope complement}. Polytope complements can be used to represent non-convex and unbounded spaces. \defn{Hyperrectangles} are convex polytopes that can be described by an upper and lower bound on each variable. The radius of a hyperrectangle is a vector $r$ containing values equal to half the interval between the upper and lower bound for each dimension. We assume that the domain of $f$ given by $\dom{f}$ is a hyperrectangle, although this domain is not included in the formulations presented here. \defn{Hypercubes} are hyperrectangles with a uniform radius. Let 
\begin{equation}
    \mathcal{X} \subseteq \mathbb{R}^n \qquad \mathcal{Y} \subseteq \mathbb{R}^m
\end{equation}
be input and output sets which we will use in our problem definitions. 
\\\\
\xhdr{Neural Verification Problem}
One approach to verifying a network is to show that an input-output property holds. Such a property can be specified as
\begin{equation}
\label{eqn:verification-problem}
    \vec{x} \in \mathcal{X} \implies \vec{y} = f(\vec{x}) \in \mathcal{Y}
\end{equation}
We will refer to any algorithm that solves verification problems as a \defn{verifier}. Different verifiers can handle different types of input and output sets. There are approaches to verification that are \defn{sound}, \defn{complete}, or both. If a sound algorithm reports that a property holds, then it must actually hold. If a complete algorithm reports that a property is violated, then it must actually be violated. An algorithm that is both sound and complete must always give a correct answer if it terminates.

\subsection{Approaches to Verification}
\label{subsec:approaches to verification}

In this section, we present several categories of verification algorithms described by \citet{liu2019algorithms}. Each approach can be extended to perform optimization as described in \cref{sec:convert verifiers}. There are four categories: reachability, optimization, \searchandreach, and \searchandopt. Although we focus on sound and complete verifiers, there are many interesting incomplete verifiers in these categories as well. 

\subsubsection*{Reachability}

Complete reachability methods compute an exact output reachable set, then use this reachable set to solve verification problems. The reachable set is found by propagating the input set through the network layer by layer. Once an exact output reachable set is found, the verification problem can be solved by checking whether the reachable set is contained within the set $\mathcal{Y}$. If the input set is a polytope or union of polytopes, then the reachable set will also be a union of polytopes \citep{xiang2017reachable}. In this case, the test for inclusion requires solving several linear programs (LPs). If the exact reachable set is contained within the output set $\mathcal{Y}$, the property holds. If it is not, then the property does not hold.   

ExactReach is a reachability method which propagates polytopes through the network \citep{xiang2017reachable}. The more recent NNV uses other set representations, for example \defn{star sets}, to greatly improve the efficiency of the propagation \citep{tran2019star,tran2020nnv,tran2020verification}.

\subsubsection*{Optimization}
Optimization methods encode the network and property as a constrained optimization problem. They typically constrain the input to be in $\mathcal{X}$ and the output to be in $\compl{\mathcal{Y}}$, representing the region outside of the output region of the property \citep{liu2019algorithms, lomuscio2017approach}. If the resulting optimization problem is feasible, then we know some input in the input space $\mathcal{X}$ can reach outside of the output space $\mathcal{Y}$, and the property must not hold. Conversely, if the optimization problem is infeasible then there must be no input in $\mathcal{X}$ that reaches outside of the output set and the property holds.

\defn{NSVerify} is an optimization method which uses a mixed integer encoding for the network and linear input and output constraints \citep{lomuscio2017approach}. \defn{MIPVerify} is another optimization method which improves on NSVerify in two ways. It adopts a tighter encoding of the ReLU and it performs a progressive bound tightening process. MIPVerify has been used to solve both output and minimum adversarial perturbation optimization problems described in \cref{subsec:neural optimization problem}.

\subsubsection*{Search}
\Searchandreach and \searchandopt methods search for a counter-example to the property by breaking the problem into a series of subproblems. The search space commonly consists of input ranges or neuron activations \citep{katz2017reluplex, wang2018efficient, wang2018formal,  liu2019algorithms, botoeva2020efficient}. The search ends immediately if it finds an input that violates the property. If a region is proven to satisfy the property, the search proceeds to the next region. If neither conclusion is reached, the region is broken down further. At each step, a reachability or optimization approach is taken to determine whether the property in the region holds or is violated.

\defn{Neurify} is an example of a \searchandreach algorithm.  It performs a type of symbolic reachability analysis called \defn{symbolic linear relaxation} to find an approximate reachable set and reason about the property for a given region \citep{wang2018efficient}. \defn{Reluplex} is an example of a \searchandopt algorithm which searches the activation space, solving a constrained linear program at each step to reason about the region \citep{katz2017reluplex}.  An extension of the Reluplex algorithm to perform optimization is described in more detail in \cref{subsec:reluplex algorithm}.

\subsection{The Reluplex Algorithm}
\label{subsec:reluplex algorithm}

The Reluplex algorithm searches for a counter-example to the property, i.e., an input $\vec{x} \in \mathcal{X}$ such that the output $\vec{y} \in \compl{\mathcal{Y}}$. The Reluplex algorithm explores the activation space by fixing ReLUs to be either inactive or active, one at a time. The search space is a binary tree, where each node represents a set of fixed ReLUs, which is equivalent to a partial activation state. Each edge leaving a node represents another ReLU becoming fixed to be active or inactive respectively. At each node, a relaxed linear feasibility problem is solved using the simplex algorithm. If a satisfying assignment is obtained, it is checked against the remaining non-linear constraints. If the obtained assignment satisfies the non-linear constraints, a counterexample has been found and the search stops. However, if the assignment violates some non-linear constraints, then Reluplex can either fix the assignment and continue solving the linear relaxation, or perform a \defn{case split}. A case split fixes the phase of a single ReLU and adds its --- now linear --- constraints to the relaxation, simplifying the problem. Reluplex judiciously explores and trims the search space until it either finds a satisfying assignment or proves that one does not exist. 

Note the similar pattern of search in Reluplex and Neurify --- both decompose the problem into smaller problems and then solve each independent feasibility problem. This search structure enables these algorithms to perform optimization, as we will show in \cref{sec:convert verifiers}.

\subsection{Neural Optimization Problem}
\label{subsec:neural optimization problem}

Neural verification problems allow us to answer yes or no questions about properties of the network. However, we would like to be able to answer qualitative questions. To that end, we define an \defn{output optimization problem} and a \defn{minimum adversarial perturbation problem}. The first consists of optimization on the output of the network subject to constraints on the input. The second consists of optimization on the input of the network subject to constraints on the output. Both can be used to answer questions that provide insight into the robustness of a system. The first problem can be used to ask questions like, ``If there is at most 5\% error in each pixel in an image used to control the steering wheel of a car, how drastically could the wheel mistakenly be turned?'' The second problem can be used to ask questions like, ``What is the smallest perturbation to my input image which would lead to me misclassifying a person as a stop sign?'' For our problems, we will consider input and output constraints given by polytopes. These can each be represented with a set of linear inequalities.

\begin{enumerate}
    \item \textbf{Output Optimization Problem}:
    We would like to find the maximum of a linear function of the output given constraints on the input. Let our objective be $g(\vec{x}) = \vec{c}^\top f(\vec{x})$, described by the user-defined parameter $\vec{c} \in \mathbb{R}^m$. More complex non-linear objectives can be approximated by augmenting the network with extra layers. For example, for a single output network with output $y$, if we wanted to maximize $y^2$ we could add on layers which approximate the function $f(y) = y^2$ and then perform verification on this augmented network. The problem formulation is
     \begin{equation}
     \label{eqn:output-opt-equation-1}
        \begin{aligned}
            & \underset{\vec{x}}{\text{maximize}} && \vec{c}^\top f(\vec{x}) \\
            & \text{subject to} 
            && \vec{x} \in \mathcal{X}
        \end{aligned}
    \end{equation} 
    with corresponding optimal value $p^*$ and optimizing input $\vec{x}^*$.
    \\
    \item \textbf{Minimum Adversarial Perturbation Problem}: We would like to find the minimum adversarial perturbation to some original input $\vec{x}_0$ that causes undesired behavior. What it means to be a small perturbation can differ between applications and algorithms, with typical distance metrics including the $L_1$, $L_2$, and $L_\infty$ norms~\citep{carlini2017provably, tjeng2017evaluating}. Typically, algorithms search for the smallest possible perturbation that causes a mistake \citep{szegedy2013intriguing, carlini2017towards, yuan2019adversarial}. In our case, we will focus on the $L_\infty$ norm for simplicity. The $L_1$ norm is also commonly used, as it can also be represented with linear constraints \citep{tjeng2017evaluating}. A perturbed input $\vec{x}$ is considered adversarial if it is in some output set $\mathcal{Y}$ which can be used to represent undesired behavior. Our problem is then
    \begin{equation}
    \label{eqn:min-adversarial-input}
        \begin{aligned}
            & \underset{\vec{x}}{\text{minimize}}
            & & \norm{\vec{x} - \vec{x}_0}_\infty \\
            & \text{subject to}
            & & f(\vec{x}) \in \mathcal{Y}
        \end{aligned}
    \end{equation}
    with corresponding optimal value $p^*$ and optimal input $\vec{x}^*$.
\end{enumerate}
These two general problems can be used to represent the two approaches for limiting the size of the perturbation and requiring adversarial behavior in the taxonomy of adversarial examples given by \citet{yuan2019adversarial}. Adversarial examples must be close to a nominal input and lead to undesirable behavior. The first problem can achieve this ``closeness'' by constraining the input set, and the undesirable behavior through the linear objective, while the second can achieve ``closeness'' by optimizing the size of the perturbation and the undesirable behavior through the output constraints.

\subsection{Approximate Methods for Optimization}

There are a variety of approximate approaches for solving neural optimization problems. We highlight several that we use to compare to \optimizer. Approximate methods provide bounds on the optima for these problems.

Many adversarial attacks provide approximate solutions to one or the other of our optimization problems \citep{yuan2019adversarial}. The Fast Gradient Sign Method (FGSM) and Projected Gradient Descent (PGD) provide approximate solutions to the output optimization problem by making use of the gradient of the objective with respect to the input \citep{goodfellow2014explaining, madry2017towards}. FGSM takes a single step in the direction of the gradient to the boundaries of the input set, while PGD takes many gradient steps, projecting into the input set after each step. LBFGS, a quasi-Newtonian optimization method \citep{zhu1997algorithm}, can also be used to find an approximate solution to the output optimization problem. We compare \optimizer~on these problems to these three methods.

\section{Strategies for Converting Verifiers into Optimizers}
\label{sec:convert verifiers}
The main observation of this paper is that many existing approaches to neural network verification can be extended to solve optimization problems. To substantiate our claim, we illustrate strategies for four major categories of verification algorithms outlined in the survey of verification methods \citep{liu2019algorithms}. As summarized in \cref{subsec:approaches to verification}, these categories are
reachability,
optimization,
search with reachability, and
search with optimization.
Each of these categories poses different advantages and challenges when being extended to support optimization. 
We describe how to modify algorithms in each of these categories, and as a proof of concept, we showcase the extension of Marabou, a state-of-the-art search and optimization verifier \citep{katz2019marabou}, and evaluate the performance of the resulting optimizer. Although in this work we only extend Marabou, there are a wide variety of other verifiers that once extended could provide a suite of optimizers each with their own disadvantages or advantages \citep{VNN20,VNN21,liu2019algorithms}.

First, however, we present how to solve optimization problems by combining a decision procedure with bisection search. This is a well-established approach that serves as a baseline for comparison with the integrated approaches that we propose.  While our focus is on complete verification procedures, we discuss extensions of incomplete approaches in \cref{sec:conclusion}. Throughout the section we consider a network $N$ represented by function $f$ with $n$ inputs, $m$ outputs, and $K$ layers.
\subsection{Optimization using Bisection Search}
\label{subsec:bisection}

A complete verifier answers yes or no to the question: ``Does $\vec{x} \in \mathcal{X}$ imply $\vec{y} \in \mathcal{Y}$?'' An optimization problem seeking to maximize some function $g(\vec{x})$ subject to either $\vec{x} \in \mathcal{X}$ or $\vec{y} \in \mathcal{Y}$ can be solved by asking a series of yes/no questions where a verification problem is constructed to represent the question, ``can the optimal value $p^*$ be greater than $d$'' for some value $d$. If we start with inital bounds on $p^*$, $\ell \leq p^* \leq u$, we can then select $d$ to perform bisection and update the bounds on the optimal value, halving the remaining search space with each step. If the answer to a query is yes, the lower bound $\ell$ is strengthened; otherwise, the upper bound $u$ is weakened. This procedure finds the optimal solution when $\ell = u$;  alternatively, it can be made to halt when a user-defined minimum optimality gap is achieved. This approach has been used to find optimal solutions \citep{julian2020validation, carlini2017provably} and often relies on bracketing techniques \citep{kochenderfer2019algorithms} to narrow in on the optimal solution. Meta strategies can be applied to further speed up the convergence, like solving multiple instances in parallel, and splitting the region into sets of disjoint intervals, rather than two halves \citep{julian2020validation}. This algorithm always terminates if the bounds are represented with floating point numbers but may not terminate if they are represented with real numbers.

\subsection{Reachability}
\label{subsec:Converting Reachability}

Reachability methods operate on sets of points rather than individual points in order to compute an exact output reachable set for a network. Sets of points are represented using abstract domains, such as polytopes or star sets, with efficient operations to propagate these domains through the layers of the network.
The verification property is of the form $\vec{x} \in \mathcal{X} \implies \vec{y} \in \mathcal{Y}$, where $\mathcal{X}$ is the input and $\mathcal{Y}$ the output set.
Reachability methods propagate abstract representations of the input set $\mathcal{X}$ through the network layers, until a reachable set $\mathcal{R}$ for the output layer is computed. The property holds if $\mathcal{R} \subseteq \mathcal{Y}$.
To reason soundly about the output set, the abstract domain needs to be either:
\begin{inparaenum}
\item exact  --- no values are lost or added during the computation, or
\item an over-approximation --- no values are lost during the computation.
\end{inparaenum}
In case of over-approximations, the subset check may fail even though the property holds. Over-approximated output reachable sets can still be used in complete methods when incorporating search in the process. The remainder of this section focuses on methods using exact representation. ExactReach \citep{xiang2017reachable} and NNV \citep{tran2020nnv} represent reachable sets using a union of convex sets --- polytopes and star sets, respectively. A star set is a polytope encoding that supports efficient computation of propagation through a network.

The reachable set can be represented as
\begin{equation}
    \mathcal{R} = \bigcup_{i=1}^k P_i,
\end{equation}
where $P_i$ are convex sets for $ i \in \{1,2,\ldots, k\}$. Therefore, there are two equivalent reachability problems to check whether the property holds: \begin{equation}
    \left(\vec{x} \in \mathcal{X} \implies \vec{y} \in \mathcal{Y}\right) \iff \left( \mathcal{R} \subseteq \mathcal{Y} \right) \iff \left(P_i \subseteq \mathcal{Y}, \; i =1, \hdots, k \right) 
\end{equation}
Checking whether $\mathcal{R} \subseteq \mathcal{Y}$ is challenging because $\mathcal{R}$ may be non-convex. 
However, we can check whether a convex set $P_i$ is contained within the polytope $\mathcal{Y}$ in polynomial time. The polytope subset problem $P \subseteq P'$, where $P,\,P'$ are polytopes and $P'$ consists of $j$ linear constraints, is answered by solving $j$ linear programs (LPs). Suppose $\mathcal{Y}$ consists of $\ell$ linear constraints, then solving a total of $k \cdot \ell$ LPs answers the verification problem in polynomial time.\footnote{Solving each LP takes polynomial time, but note that $k$ may be exponentially large compared to the input representation. This exponential growth in the number of output sets is a challenge for both reachability verifiers and optimizers.}


Given the basic overview of reachability techniques, how do we solve an output optimization problem using the exact reachable set $\mathcal{R}$? Let $\vec{c}^\top y$ with $c \in \mathbb{R}^m$ be the objective function.
The optimal value $p^*$ can be expressed in several ways:
\begin{align}
    p^* &= \max_{\vec{x} \in \mathcal{X}} \vec{c}^\top f(\vec{x}) \\
    &= \max_{\vec{y} \in \mathcal{R}} \vec{c}^\top\vec{y} \label{eq:reachopt-nonconvex} \\
    &= \max_{i \in \{1, 2, \hdots, k\}} \max_{y \in P_i} \vec{c}^\top \vec{y} \label{eq:reachopt-convex}
\end{align}
There are two ways to answer the optimizing query using reachability:
\begin{inparaenum}
\item using the non-convex set $\mathcal{R}$ as shown in \cref{eq:reachopt-nonconvex}, or
\item using the set of convex sets $P_i,\, i \in \{1,2, \ldots, k\}$ as shown in \cref{eq:reachopt-convex}.
\end{inparaenum} In the latter case, the inner maximization of the final expression $\max_{y \in P_i} \vec{c}^\top \vec{y}$ has a linear objective and linear constraints, so it is an LP. Consequently, solving an LP for each index $i$ yields an exact value of $p^*$.

Note that, whereas solving the subset problem uses $\ell$ LPs per polytope, a single LP per polytope $P_i$ is sufficient to find the optimal value. The maximum of the local maxima for each $P_i$ yields the global optimum. Given the reduction in the number of LP problems solved, there are likely to be cases where the optimization problem is more efficient than the verification problem with the same input set.  Understanding when this is the case is an interesting area for future investigation.  We also expect this approach to outperform the bisection approach, as the latter requires many more calls to the verifier (albeit with smaller input sets each time).

Exact reachability-based methods can readily be extended to solve output optimization problems. This means that ExactReach and NNV with minimal modifications could be applied to an output optimization problem; the only change required is replacing the polytope subset check with a single optimizing LP call. 

Although the approach outlined above guarantees an optimal value, it does not provide a method to find the corresponding optimizing input. Doing so would require tracking additional information linking input regions to the $P_i$ polytopes. Specific implementations of reachability-based approaches may or may not track this information. Those that do typically maintain a one-to-one correspondence between polytopes in the input space and output space \citep{tran2020nnv,vincent2020reachable}. 
Tracking this correspondence provides an additional benefit: we can find the exact input set that maps to a specified output polytope. This is accomplished by mapping the intersection of the output polytope with the exact reachable set back to the input space as a union of polytopes \citep{vincent2020reachable}.
We can use this to solve the minimum adversarial perturbation problem. We first find the pre-image of the output set $\mathcal{Y}$ in the input space. We then apply the same approach to analyzing this union of polytopes as we did with the output optimization problem: we can iterate one by one through the polytopes, solving a convex program at each step. In this case the objective is to minimize $\norm{\vec{x} - \vec{x}_0}_\infty$ over each polytope. As a result, we can extend exact reachability-based methods that maintain a correspondence between the input and output sets to solve the minimum adversarial perturbation problem.




\subsection{Optimization}
\label{subsec:Converting Optimization}

In this section, we discuss extension of optimization-based verification approaches to solve optimization problems. This is a fairly direct modification from a theoretical standpoint.
Optimization approaches, such as NSVerify and MIPVerify \citep{lomuscio2017approach, tjeng2017evaluating}, exactly encode the network as a mixed integer program (MIP). The input set $\mathcal{X}$ and the complement of the output set $\mathcal{Y}$ are added as linear constraints to the MIP, resulting in a feasibility problem
\begin{equation}
	\label{eq:mip}
    \begin{aligned}
    & \underset{\vec{x}}{\text{maximize}} && 0 \\
    & \text{subject to} && \vec{x} \in \mathcal{X} \\
    & && \vec{y} = f(\vec{x}) \\
    & && \vec{y} \in \compl{\mathcal{Y}}
    \end{aligned}
\end{equation}

The original property 
$
	\vec{x} \in \mathcal{X} \implies \vec{y} \in \mathcal{Y} 
$
does not hold if and only if the MIP is feasible. These optimization-based approaches are sound and complete. 

As the problem statement in \cref{eq:mip} suggests, extending this formulation to support optimization problems only requires the embedding of a goal function. Indeed, MIPVerify uses this approach to solve these classes of problems \citep{tjeng2017evaluating}.

For output optimization problems, the output constraint is removed and an objective is added to the mixed integer program, resulting in
\begin{equation}
    \begin{aligned}
    & \underset{\vec{x}}{\text{maximize}} &&  \vec{c}^\top \vec{y}\\
    & \text{subject to} && \vec{x} \in \mathcal{X} \\
    & && \vec{y} = f(\vec{x})
    \end{aligned}
\end{equation}
The input set and network constraints remain unchanged. This is still an MIP, since we assume a linear output objective. Most MIP solvers readily admit MIPs with an objective, requiring no further modification in practice.

The process is similar for encoding a minimum adversarial perturbation problem described by output set $\mathcal{Y}$ and original point $\vec{x}_0$. In this case, an objective is added for the input and the input constraint is removed, while the network and output constraints remain unchanged. Alternatively, we can also replace the output constraint with $ \vec{y} \in \mathcal{Y'}$, where $\mathcal{Y'}$ represents a target set of adversarial behaviors, resulting in
\begin{equation}
    \begin{aligned}
    & \underset{\vec{x}}{\text{minimize}}
    & & \norm{\vec{x} - \vec{x}_0}_\infty \\
    & \text{subject to} && \vec{y} = f(\vec{x}) \\
    & && \vec{y} \in \mathcal{Y'}
    \end{aligned}
\end{equation}



As shown in this section, optimization-based verifiers need an appropriate objective added in order to solve output optimization or minimum adversarial input problems. This approach translates directly to \searchandopt solvers as well. In practice, this requires minimal modification to the source in order to solve a whole new class of problems.


\subsection{Search}
\label{subsec:converting search}

Search-based approaches break down the space into smaller regions, typically by constraining the input or the activation space \citep{liu2019algorithms}. \Cref{alg:search verification} gives the pseudocode for search-based verification.  At each search state, a procedure \Call{Violated}{$S$, $P$} is invoked. This procedure takes as input a state in the search space $S$ and problem $P$ and returns one of the following three outputs:
\begin{enumerate}
    \item a status of \violated and an assignment that violates the property;
    \item a status of  \holds, indicating that the property holds in this region;
    \item a status of \unknown, indicating that it is unknown whether the property holds in this region.
\end{enumerate}
If a violating assignment is found, the search returns the discovered solution. If the answer is inconclusive, the search state is decomposed into multiple smaller --- i.e., further constrained --- states and the search continues. If the property holds in a state, the search proceeds with the next unexplored state. If all states have been explored, the search procedure determines that the property holds.

\begin{algorithm}
    \caption{Search-based verification}
    \label{alg:search verification}
    \begin{algorithmic}[1]
        \Function{Check}{$S$, $P$}
            \State states $\gets$ [$S$]
            \While{ states is not empty }
                \State state $\gets$ states.dequeue();
                \State status, assignment $\gets$ \Call{Violated}{state, P}
                \If{status = \violated}
                    \State \textbf{return} Counter-example( assignment )
                \ElsIf{status = \unknown}
                    \State states.enqueue(\Call{Split}{state, P})
                \EndIf
                \Comment If the status is \holds the search continues
            \EndWhile
            \State \textbf{return} PropertyHolds
        \EndFunction
    \end{algorithmic}
\end{algorithm}


\begin{algorithm}
    \caption{Search-based optimization, branch and bound}
    \label{alg:search opt}
    \begin{algorithmic}[1]
        \Function{Optimize}{$S$, $P$}
            \State states $\gets$ [$S$]
            \State optSoFar $\gets$ None 
            \State $x$ $\gets$ None
            \While{ states is not empty }
                \State state $\gets$ states.dequeue()
                \State status, val, assignment $\gets$ \Call{OptimumForRegion}{state, P, optSoFar}
                \If{status = Unknown}
                   \State states.enqueue(\Call{Split}{state, P})
                \ElsIf{ status = Optimal and val $>$ optSoFar
                }
                        \State optSoFar $\gets$ val
                        \State $x$ $\gets$ val
                \EndIf
                \Comment If the status is WorseThanOpt the search continues 
            \EndWhile
            \State \textbf{return} optSoFar, $x$
        \EndFunction
    \end{algorithmic}
\end{algorithm}

\Cref{alg:search opt} extends the search-based verification approach to solve optimization problems instead. $S$ is still a state, and $P$ is an optimization problem which includes an objective function and constraints. The \Call{Violated}{} procedure is replaced by an \Call{OptimumForRegion}{state, $P$, optSoFar} optimizing procedure. This can return one of the following three outputs:
\begin{enumerate}
    \item a status of \worsethanopt which indicates that the optimal value in this region is less than the true optimum. The function can make use of optSoFar to make this assertion.
    \item a status of \unknown, indicating that it is unknown whether the true optimum could be contained in this region.
    \item a status of \optimal, an objective value guaranteed to be the optimum in this region, and an assignment which achieves this objective value. This indicates that it has found a true global optimum.
\end{enumerate}
In order to be terminating it also must be guaranteed to find the optimal, and not return \unknown, once the region has been split enough times. One example of an implementation of \Call{OptimumForRegion}{state, $P$, optSoFar} would be to solve a relatively tractable relaxed problem, providing an upper bound on the solution. If this upper bound is less than optSoFar, it can return \worsethanopt since the upper bound being lower than an objective value that has already been achieved will guarantee the region cannot achieve the optimum. If the assignment from solving the relaxed problem achieves the upper bound, it can return the objective value and assignment along with a status of \optimal. Otherwise, it can return \unknown. The way that the problem is relaxed and the upper bound is computed differs between \searchandreach and \searchandopt strategies.

\Searchandreach verification methods will typically use approximate reachability methods in place of \Call{Violated}{} \citep{wang2018formal, wang2018efficient}. To convert these verifiers into optimizers, their \Call{Violated}{} procedure can be modiified to implement \Call{OptimumForRegion}{} in a similar way to the extension of pure reachability-based methods covered in \cref{subsec:Converting Reachability}. This could be applied to extend verifiers such as Neurify and ReluVal, which compute an approximate reachable set at each step \citep{ wang2018efficient, wang2018formal}. Similarly, \searchandopt~methods use constrained optimization problems to implement \Call{Violated}{} \citep{katz2017reluplex, botoeva2020efficient}, which naturally extend to optimization as discussed in \cref{subsec:Converting Optimization}. In our extension of Reluplex we take this approach, solving a relaxed problem that can provide an upper bound on the true optimal value for a partial activation state \citep{katz2017reluplex}. This is described in more detail in \cref{sec:modifying marabou}.

In practice, the search structure largely remains the same as verification, with the exception that finding a counter-example does not end the search. Instead, such intermediate values are used to facilitate the branch-and-bound strategy.

\section{Extending Reluplex to Solve Optimization Problems}
\label{sec:modifying marabou}

Reluplex is a \searchandopt technique. As such, we can make use of the strategy presented in \cref{subsec:converting search} to convert Reluplex from a verifier into an optimizer. First, we define the \Call{Split}{$S$, $P$} and \Call{Violated}{$S$, $P$} functions used in the original Reluplex; then, we will discuss how to convert the \Call{Violated}{$S$, $P$} function into a corresponding \Call{OptimumForRegion}{$S$, $P$, optSoFar} function.

In order to solve a verification problem with a \searchandopt technique, we need to define two functions: \Call{Split}{$S$, $P$} and \Call{Violated}{$S$, $P$}. The $\Call{Split}{}$ function is meant to take a state and break it down into two (or more) states which will be easier to solve. This guides the search process. Reluplex searches over the activation states of the network. It does this in an incremental fashion, starting at a state where no ReLUs are fixed and then proceeding to fix ReLUs one by one during the search. Here, the state $S$ will be a partial activation state. The implementation of $\Call{Split}{}$ will choose a node $(i, j)$ that is undetermined in the current state and return two new states: the first state will be the current state with the additional node $(i, j)$ fixed to be active, and the second state will be the same except with node $(i, j)$ fixed to be inactive. There are a variety of possible strategies to pick this node to fix, including choosing the earliest unfixed ReLU or choosing the ReLU with the largest violation. Bounds on the variables may fix some ReLUs to be a certain phase, in which case those ReLUs will not need to be split.

The function \Call{Violated}{$S$, $P$} should reason about whether a property $P$ holds for the partial activation state $S$. Reluplex accomplishes this by relaxing each undetermined ReLU from $z = max(0, \hat{z})$ to $z \ge 0 \land  z \ge \hat{z}$. A variety of other relaxations for a ReLU are possible \citep{liu2019algorithms}. Using this relaxation means that these undetermined ReLUs can each be written with two linear constraints, allowing the relaxed feasibility problem to be an LP. Recall that $\mathcal{L}$ represents the indices of ReLU layers, $\mathcal{I}$ represents the indices of identity layers, $\mathcal{A}$ gives the set of active nodes in the activation state, $\mathcal{N}$ gives the set of inactive nodes for the activation state, and $\mathcal{U}$ gives the set of undetermined nodes for the activation state. We also assume we have upper and lower bounds on each variable denoted as $\hat{U}_{(i, j)}$ and $\hat{L}_{(i, j)}$ for pre-activation variables, and $U_{i, j}$ and $L_{i, j}$ for post activation variables for each node $(i, j)$. With all of these variables defined, the relaxed feasibility problem can now be formulated as
\begin{equation}
\label{eq:violated defn}
    \begin{aligned}
        & \underset{\vec{x}, \hat{\vec{z}}, \vec{z}}{\text{maximize}} && 0 \\
        & \text{subject to} && \vec{x} \in \mathcal{X} \\
        &&& \vec{z}_K \in \compl{\mathcal{Y}} \\
        &&& \hat{\vec{z}}_{i+1} = \mat{W}_i\vec{z}_i + \vec{b}_i, \; i = 1,\hdots,K-1 \\
        &&& \vec{z}_{i} = \hat{\vec{z}}_{i}, \; i \in \mathcal{I} \\
        &&& z_{i,j} = \hat{z}_{i,j}, \; (i, j) \in \hat{\mathcal{A}} \\
        &&& z_{i,j} = 0, \; (i,j) \in \hat{\mathcal{N}} \\
        &&& z_{i, j} \ge z_{i,j} \; (i, j) \in \mathcal{U} \\
        &&& z_{i, j} \ge 0 \; (i, j) \in \mathcal{U} \\
        &&&  \hat{L}_{i, j} \le \hat{z}_{i,j} \le \hat{U}_{i,j} \\
        &&& L_{i,j} \le z_{i,j} \le U_{i,j}
    \end{aligned}
\end{equation}
Recall that the complement of the output set is applied as a constraint, representing the search for a counter-example.

Since this problem is a relaxation, if the problem is infeasible then the exact problem must also be infeasible. If the problem is feasible, then the satisfying assignment from the LP can be checked to see if it is consistent with the exact problem.  If it is, then we have found a counter-example. 
If not, then the feasibility of the exact problem remains unknown. These observations can be used to construct the \Call{Violated}{} function.

To extend the Reluplex algorithm to perform optimization, we need to extend the functionality of the \Call{Violated}{$S$, $P$} procedure to create a \Call{OptimumForRegion}{$S$, $P$} procedure. We will convert the feasibility problem used to implement \Call{Violated}{$S$, $P$}, \cref{eq:violated defn} into a linear program which will provide us an upper bound on the objective. This requires adding an objective to the problem being solved just like with purely optimization approaches in \cref{subsec:Converting Optimization}. If we have an objective function $g(\vec{x})$ and assume without loss of generality the objective is being maximized, we arrive at the formulation
\begin{equation}
\label{eq:relaxed with obj}
    \begin{aligned}
        & \underset{\vec{x}, \hat{\vec{z}}, \vec{z}}{\text{maximize}} && g(\vec{x}) \\
        & \text{subject to} && \vec{x} \in \mathcal{X} \\
        &&& \vec{z}_K \in \compl{\mathcal{Y}} \\
        &&& \hat{\vec{z}}_{i+1} = \mat{W}_i\vec{z}_i + \vec{b}_i, \; i = 1,\hdots,K-1 \\
        &&& \vec{z}_{i} = \hat{\vec{z}}_{i}, \; i \in \mathcal{I} \\
        &&& z_{i,j} = \hat{z}_{i,j}, \; (i, j) \in \hat{\mathcal{A}} \\
        &&& z_{i,j} = 0, \; (i,j) \in \hat{\mathcal{N}} \\
        &&& z_{i, j} \ge \vec{z}_{i,j} \; (i, j) \in \mathcal{U} \\
        &&& z_{i, j} \ge 0 \; (i, j) \in \mathcal{U} \\
        &&&  \hat{L}_{i, j} \le \hat{z}_{i,j} \le \hat{U}_{i,j} \\
        &&& L_{i,j} \le z_{i,j} \le U_{i,j}
    \end{aligned}
\end{equation}
The objective function, input set, and output set can be used to represent output optimization problems and minimum input perturbation problems. The resulting optimal value $p^*$ from this LP provides an upper bound on the true optimal value for this activation state since the LP comes from a relaxation of the exact optimization problem for this region. 

In order to implement \Call{OptimumForRegion}{$S$, $P$, optSoFar} we first solve the LP in \cref{eq:relaxed with obj}. If the optimal value, representing an upper bound on the objective, is lower than optSoFar, then status \worsethanopt is returned. Otherwise,  the procedure checks whether the found upper bound can actually be achieved --- this is done by passing the corresponding input value through the network, or checking whether all ReLU constraints are met. If it does, it is the optimum value for the region, and status \optimal with this optimal objective value and assignment is returned. Otherwise, status \unknown is returned, indicating that the search should continue in sub-regions. The relaxed LP becomes exact when all activation states are defined; therefore, this approach is guaranteed to find the optimum in a fully defined activation state, i.e., in a leaf node of the binary search tree representing the activation space. This ensures that the procedure will return an optimum in every leaf node, ensuring termination of the algorithm. 
\subsection{Example}
\label{subsec: tree search example}

\begin{figure}
\centering
\begin{tikzpicture}[font=\footnotesize,scale=0.75]
			\matrix [ampersand replacement=\&,row sep=0.7cm,column sep=0.3cm] {
			\node {No ReLUs fixed};	\& \& \& \node [circle,draw=black] (a) {$20$};\\
			\node {1 ReLU fixed};	\& \& \node [circle,draw=black] (b) {$17$}; \& \& \node [circle,draw=black] (c) {$-\infty$};\\
			\node {2 ReLUs fixed};	\& \node [circle,draw=black,fill=black!20] (d) {$9$}; \& \& \node [circle,draw=black] (e) {$7$};\\
				};
			\draw [->] (a) -- (b);
			\draw [->] (a) -- (c);
			\draw [->] (b) -- (d);
			\draw [->] (b) -- (e);
\end{tikzpicture}
\caption{ Example search tree with LP objective values }
\label{tree_figure}
\end{figure}
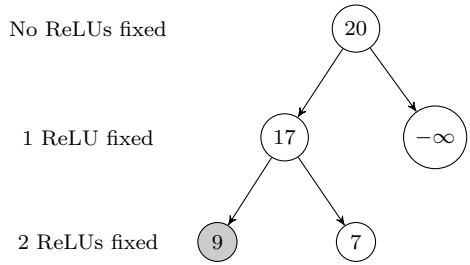

The optimization process is illustrated in \cref{tree_figure}. Each node contains the optimum value of the relaxed LP associated with that partial activation state. Gray shading indicates that the optimum of the relaxed LP satisfies all linear and ReLU constraints and so is a valid candidate for global optimum. This figure shows that the algorithm performs two splits and then finds a solution that satisfies both the linear and nonlinear constraints (the gray shaded node). The maximum value found is 9.  This can then be used to trim the branch to its right, where the best value found is only 7. The rightmost node is found to be infeasible, completing the search.  The optimal value is 9.


\if{false}
\Chris{This feels a bit out of place but I'm not sure where it would be best}
\subsection{Preprocessing and sharing modules between solvers}
\label{subsec:preprocessing}

We found that incorporating MIPVerify's approach to preprocessing into \optimizer had a significant impact on its performance, especially on perception networks. This is discussed in \cref{subsubsec:cost of preprocessing}. 

MIPVerify employs a progressive bound tightening preprocessing strategy where it (i) computes bounds with interval arithmetic, (ii) formulates a relaxed LP for each node to obtain tighter bounds than found with interval arithmetic, and (iii) formulates a mixed integer exact encoding of the network and solves for lower and upper bounds at each node with a timeout for each bound \citep{tjeng2017evaluating}. You can choose between these different levels of preprocessing, allowing for flexibility in the tradeoff between preprocessing time and solve time. We were able to incorporate a similar preprocessing approach into \optimizer~which already employed a host of different bound tightening techniques. The approach we introduced to \optimizer~optimizes on both the pre-activation and post-activations bounds separately for each node, with a timeout for each. More exploration is needed to determine whether this is beneficial. This process of borrowing a module from MIPVerify suggests that modularizing and combining components of different algorithms may be an interesting area to explore more deeply.
\fi

\section{Experiments and Results}
\label{sec:results}
In this section, we evaluate the performance of \optimizer~on a diverse set of optimization queries, including safety properties of control systems and robustness properties of perception models.

\subsection{Implementation and Experimental Setup}

We extend Marabou \citep{katz2019marabou}, an open-source neural network verification tool implementing the Reluplex algorithm, to support solving optimization queries, using the method described in \cref{subsec:converting search} and \cref{sec:modifying marabou}. Marabou integrates the symbolic bound tightening techniques introduced in \cite{wang2018formal} and a modified version of the progressive bound tightening preprocessing pass introduced in \cite{tjeng2017evaluating}.%
\footnote{Marabou also later integrated the DeepPoly analysis from \citet{singh2019abstract}, 
which can derive tighter bounds then the symbolic bound tightening technique. This experimental evaluation was conducted before DeepPoly was made
available in Marabou.}
Note that while the tool supports parallelism in both preprocessing and solving \citep{wu2020parallelization}, the results here do not make use of any parallelism.  Integrating our optimizing extension with the parallel features of Marabou is a promising avenue for future work.

We refer to the optimizing extension of Marabou as \optimizer. Given an optimization query and a timeout, \optimizer~returns \textsc{Infeasible}, \textsc{Timeout}, or the variable assignment for the optimal solution. We compare MIPVerify \citep{tjeng2017evaluating} to \optimizer, as it can directly solve the same set of benchmarks. By extending other verifiers according to our framework, for example those described in the survey by \citet{liu2019algorithms} or those in the International Verification of Neural Networks Competitions \citep{VNN20,VNN21}, it may be possible to create a stronger tool to compare against than MIPVerify. 

\if{false}
For each optimization query, we turn on both symbolic bound tightening and the progressive bound tightening preprocessing pass. The preprocessing pass in Marabou computes bounds for both the input variable and the output variable of each ReLU. In comparison, MIPVerify computes the bounds of the input variables with an MIP solver then computes the bounds on the output variables from the input variables. This means the preprocessing pass of Marabou could take up to twice as long as MIPVerify with the same per-query timeout.
\fi

As an additional baseline, we implement a binary search to solve optimization problems with a series of calls to the Marabou verifier and the approach described in \cref{subsec:bisection}. We refer to this optimizer as MarabouBin. For output optimization queries, we first perform calls to the verifier to find an upper bound for the objective, and then begin the binary search. For minimum adversarial input queries we begin with an upper and lower bound and can thus immediately start the binary search in the middle of that interval. In all cases, we continue narrowing the interval until an optimality gap less than $10^{-4}$ is achieved. Since the minimum adversarial input queries chosen have the potential to be unsatisfiable (there is no adversarial input in the given region), the binary search will return that the query is unsatisfiable if it does not find an adversarial example to within $10^{-4}$ of the border of the region. We choose to start in the middle instead of first checking whether the full region is satisfiable, as we expect many queries to be satisfiable and queries that includes the full input region to be quite expensive. By starting in the middle, we potentially avoid ever needing to consider substantial portions of the full input region if we find an adversarial example early enough. This is a design choice for MarabouBin which we expect to produce different times for satisfiable and unsatisfiable queries than if we had first checked satisfiability.

MIPVerify tightens the bounds of the input to each neuron with an MIP-solver and then solves the preprocessed queries using a Mixed-integer encoding of the problem. Marabou employs a similar preprocessing pass, except that the tightening is done on both the input variable and the output variable of each neuron. The timeout per preprocessing query for both \optimizer~and MIPVerify is 1 second, while that for MarabouBin is 0.5 seconds. We obtain these values by performing a grid search of this parameter for each solver on a subset of the benchmarks.

In addition to complete methods, we also compare against three approximate approaches, including  Projected  GradientDescent (PGD), Limited-memory BFGS (LBFGS), and the Fast Gradient SignedMethod (FGSM).

\subsection{Benchmarks}
The benchmark sets consist of network-query pairs, with networks from three different application domains: aircraft collision avoidance (ACAS Xu), aircraft localization (TinyTaxiNet), and digit recognition (MNIST). Optimization queries include both output optimization problems and minimum adversarial perturbation problems.

\medskip 

\xhdr{ACAS Xu Family} The ACAS Xu family of benchmarks, introduced in \cite{katz2017reluplex}, implements a prototype aircraft collision avoidance system --- advising course corrections based on the relative positions of two aircraft. The system consists of 45 fully-connected feed-forward neural networks, each with 6 hidden layers and 50 ReLU nodes per layer. Each network uses 5 inputs to describe the encounter geometry and produces 5 outputs --- the predicted cost of following each action. The system chooses the action with the lowest cost as the advisory. We consider both output optimization queries and minimum adversarial perturbation queries on the 45 networks. 

For output optimization queries, the objective is maximizing $y_{real} - y_{adv}$, where $y_{real}$ is the expected output in the given input region, and $y_{adv}$ is an adversarial output. For minimum adversarial perturbation queries, the objective is to minimize the perturbation on one input dimension that would result in an adversarial output.
The input regions used in the output optimization queries are from properties $1$--$4$ in \citet{katz2017reluplex}, which apply to all $45$ networks. 
To construct the minimum adversarial perturbation queries we adopt the input and output constraints from property $2$. We then consider perturbations on a single input at a time. We set the objective for each query to be to minimize the perturbation from the center for a single input dimension. This results in $5$ distinct queries per network, one for each input dimension.
In total, this yields $180$ ($45 \times 4$) output optimization queries and 225 ($45 \times 5$) input optimization queries.
\\\\
\xhdr{TinyTaxiNet} The TinyTaxiNet family of benchmarks consist of perception networks that predict the aircraft position on the taxiway relative to the center-line. The output is used by a controller that adjusts the trajectory to correct the position of the aircraft. The input to the network is a gray-scale image compressed to 16x8 pixels, with values ranging between [0,1]. The networks produce two outputs: the lateral distance to the runway center-line and the heading angle with respect to the center-line. We evaluated on three network architectures, each consisting of one convolution layer and 2 feed-forward layers. The networks have a total of 32, 64, and 128 ReLUs, respectively. 

We consider the problem of output optimization on these benchmarks. The task is to maximize the predicted lateral distance to the runway center-line. The input region is a hyper-cube parametrized by the centroid and the radius. For each network, we generate 60 such queries, with centroids randomly sampled from the training data and the radius sampled evenly from the set $\{0.04, 0.08, 0.016\}$. 
\\\\
\xhdr{MNIST}
 We also evaluated \optimizer~on four fully-connected feed-forward networks trained on the MNIST dataset of handwritten digits. Each network has 784 inputs (representing a grey-scale image) with value range
[0,1] and 10 outputs (each representing a digit). We trained 4 models. MNIST1 and MNIST2 consist of 10 layers each, with 10 and 20 ReLU nodes per layer respectively. MNIST3 and MNIST4 consist of 20 layers each, with 20 and 40 ReLU nodes per layer respectively. The networks have approximately \num{95}\% accuracy on the MNIST test set. 
The range of widths and depths of these networks, with a maximum of 800 ReLUs, allows for queries of varying computational difficulty. State-of-the-art approximate verification techniques on local robustness queries can often scale to much larger networks, even some with hundreds of thousands of nodes \citep{muller2020neural}.

 
 We consider minimum adversarial perturbation queries on the MNIST networks. The task is to minimize the $L_\infty$ perturbation on the input image that results in an adversarial output. We generate $50$ such queries for each network by randomly choosing training images, $5$ from each class, and corresponding target labels. We set the radius of the input region to be 0.05, which corresponds to a maximal perturbation of 13 pixel-values. These queries will be infeasible if there is no adversarial input within the input region. The extended verifiers we propose can handle infeasible queries.
 
 In summary, the full benchmark set consists of $225$ minimum adversarial perturbation queries on ACAS Xu networks (\textbf{ACAS In.}), 180 output optimization queries on the ACAS Xu networks (\textbf{ACAS Out.}), $180$ output optimization queries on the TinyTaxiNets (\textbf{Taxi Out.}), and $200$ minimum adversarial perturbation queries on the MNIST networks (\textbf{MNIST Out.}).

\subsection{Experimental Evaluation}

In this section, we present results of the following experiments:
\begin{enumerate}
    \item Evaluation of the runtime performance of \optimizer and MarabouBin on the three benchmark sets. We compare against MIPVerify, a state-of-the-art solver, on the same benchmarks.
    \item Comparison of the objective values found by approximate methods with the true optimums found by exact methods.
\end{enumerate}

We run all experiments on a cluster equipped with Intel Xeon E5-2620 v4 cpus running Ubuntu 16.04. One processor with 8GB RAM was allocated for each job, and each job is given a 2-hour CPU timeout. 

\subsection{Runtime Evalution}


We ran \optimizer, MarabouBin, and MIPVerify on all benchmarks. The number of solved instances and total runtime of the solved instances are shown in \cref{fig:results}. For each benchmark set, we highlight the solver that solves the most instances. 
\\
\begin{table}[t]
\setlength\tabcolsep{2pt}
\centering		
\resizebox{1\columnwidth}{!}{ \renewcommand{\arraystretch}{1.1}
\hspace*{-0.2cm}
\begin{tabular}{@{}lccccccccc@{}}
\toprule
& \multirow{3}{*}{\vspace*{8pt}\textbf{Solver}}&\multicolumn{2}{c}{\textbf{MNIST In.} (200) }&\multicolumn{2}{c}{\textbf{ACAS In.} ( 225)}&\multicolumn{2}{c}{\textbf{Taxi Out.} (180)} &\multicolumn{2}{c}{\textbf{ACAS Out.} (180)}\\
\cmidrule(lr){3-4} \cmidrule(lr){5-6} \cmidrule(lr){7-8} \cmidrule(lr){9-10}

& & \multicolumn{1}{c}{\#Solved}& \multicolumn{1}{c}{Time} & \multicolumn{1}{c}{\#Solved}& \multicolumn{1}{c}{Time}  & \multicolumn{1}{c}{\#Solved}& \multicolumn{1}{c}{Time}
 & \multicolumn{1}{c}{\#Solved}& \multicolumn{1}{c}{Time}\\ 
 \cmidrule{2-10}
& {\bf MIPVerify} & \bf{61} & 99127 & 138 & 300558 & \bf{180} & 3150 & 90 & 1169.4 \\ 
& {\bf MarabouBin} & 29 & 53301 & 123 & 504615 & 120 & 20237 & 82 & 107422 \\
& {\bf \optimizer} & 14 & 13209 & \bf{174} & 226431 & 126 & 6144 & \bf{97} & 61864 \\
\bottomrule
\end{tabular}}
\caption{Number of solved instances and runtime in seconds. \label{fig:results}}
\end{table}

\begin{figure}[ht]
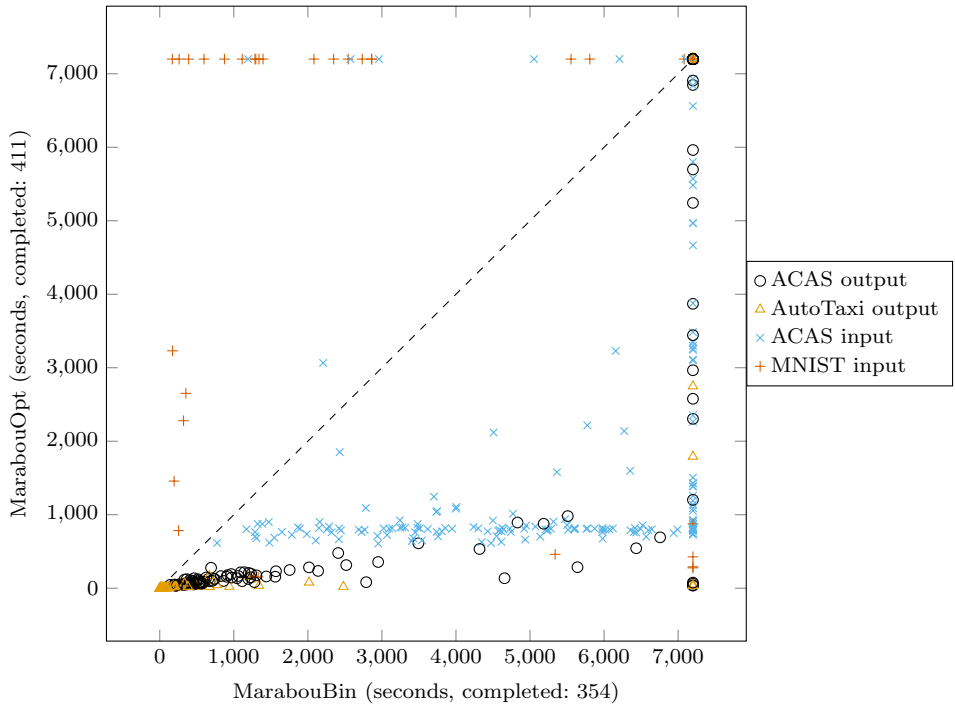

\begin{center}
%
\end{center}
\caption{Runtime comparison of \optimizer and MarabouBin.}
\label{fig:opt-vs-bin}
\end{figure}

\begin{figure}[ht]
\begin{center}
%

\end{center}
\caption{Runtime comparison of \optimizer and MIPVerify.}
\label{fig:opt-vs-mip}
\end{figure}

\begin{figure}[ht]
\centering
\includegraphics[width=10cm]{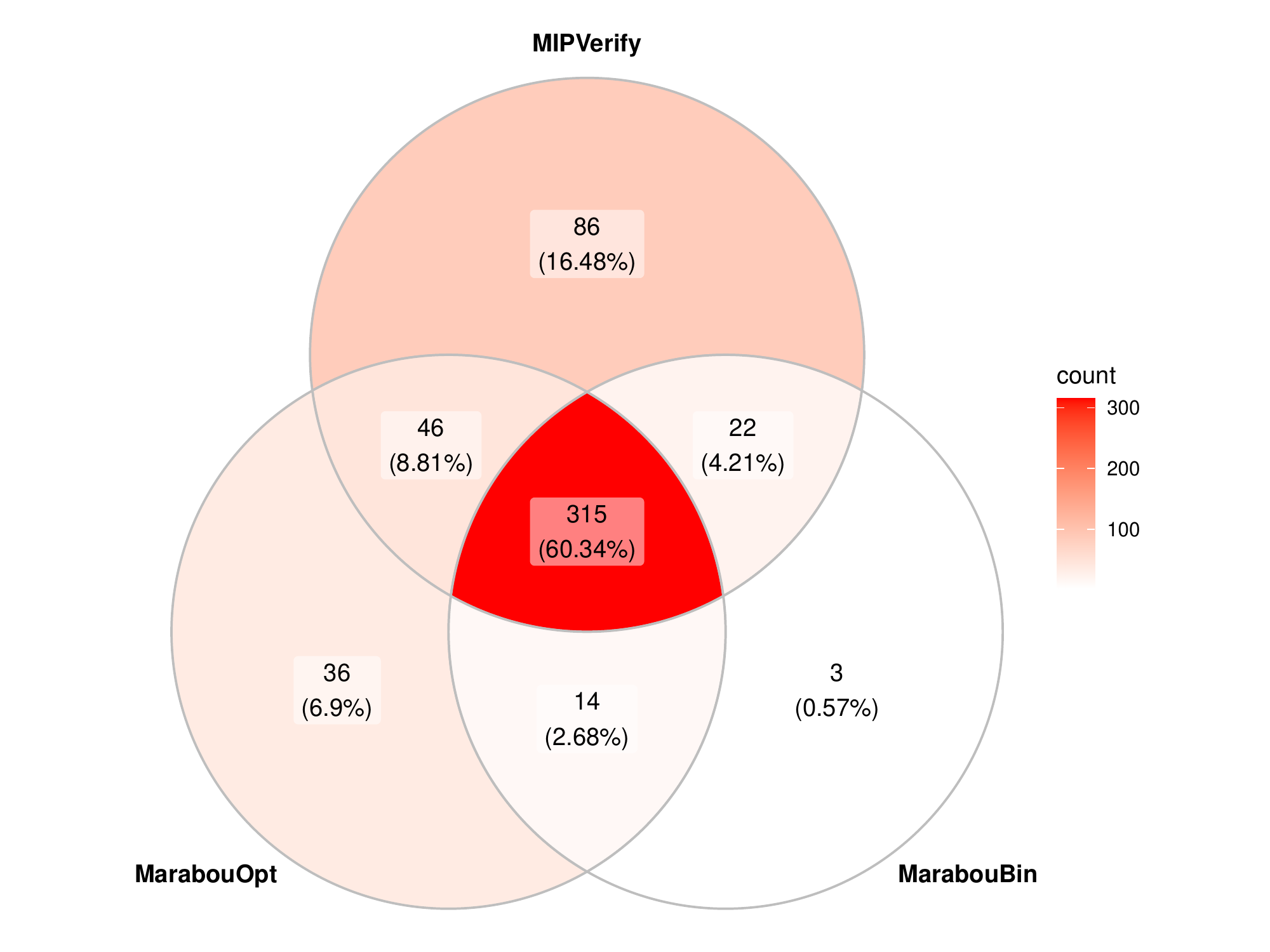}%
\caption{Number of commonly and uniquely solved benchmarks. \label{fig:venn1}}
\end{figure}

\xhdr{MarabouOpt versus MarabouBin}
As shown by \cref{fig:results}, \optimizer outperforms MarabouBin on three of the four benchmark sets, showing that 
our integrated optimization extension of the Marabou verifier is overall more effective than the black-box approach using bisection search. 
\Cref{fig:opt-vs-bin} shows a scatter plot of the runtime of the two solvers on all benchmarks. Points with value 7200 on the $x$  and $y$ axes denote timeout of the tool on the respective axis.
\Cref{fig:opt-vs-bin} shows that \optimizer significantly outperforms MarabouBin evidenced by the concentration of points below the dashed diagonal line. Surprisingly, \Cref{fig:opt-vs-bin} shows that MarabouBin outperforms \optimizer on some of the MNIST and ACAS input queries, which merits further investigation.
\\\\
\xhdr{MarabouOpt versus MIPVerify}
On the other hand, \optimizer and MIPVerify show strengths on different benchmark sets. While MIPVerify outperforms \optimizer~on the MNIST and TinyTaxiNet benchmark sets, \optimizer~solves more instances of both input and output optimization queries on the ACAS Xu networks. The complementary nature of the two solvers is further illustrated by \cref{fig:opt-vs-mip}.
The cluster of queries on the bottom left boundary with $x$ less than \num{300} and $y$ up to \num{3000} suggests that a subset of output optimization problems on the ACAS Xu benchmarks can be quickly resolved by MIPVerify while taking non-trivial time for \optimizer. On the other hand, the points in the right region with $x$ equal to \num{7200} suggest that \optimizer~is able to solve some of the harder ACAS Xu queries that MIPVerify cannot handle. The horizontal cluster with $y$ around \num{900} suggests that a number of minimum adversarial perturbation benchmarks on ACAS Xu networks take Marabou around 900 seconds to solve. This is because many of those queries are quickly resolved by \optimizer~after the preprocessing pass.
\\\\
\xhdr{Comparing solved instances for all solvers}
Figure \ref{fig:venn1} shows the number of commonly and uniquely solved benchmarks by the three solvers. There are 86 instances that can be uniquely solved by MIPVerify, and 36 instances that can be uniquely solved by MarabouOpt. The two solvers combined can cover over 99\% of the solved instances. Interestingly, there are 3 instances that only MarabouBin can solve. Upon closer examination, they are all minimum adversarial input queries. 

\begin{figure}
\begin{minipage}[t]{0.45\textwidth}
\includegraphics[width=5.5cm]{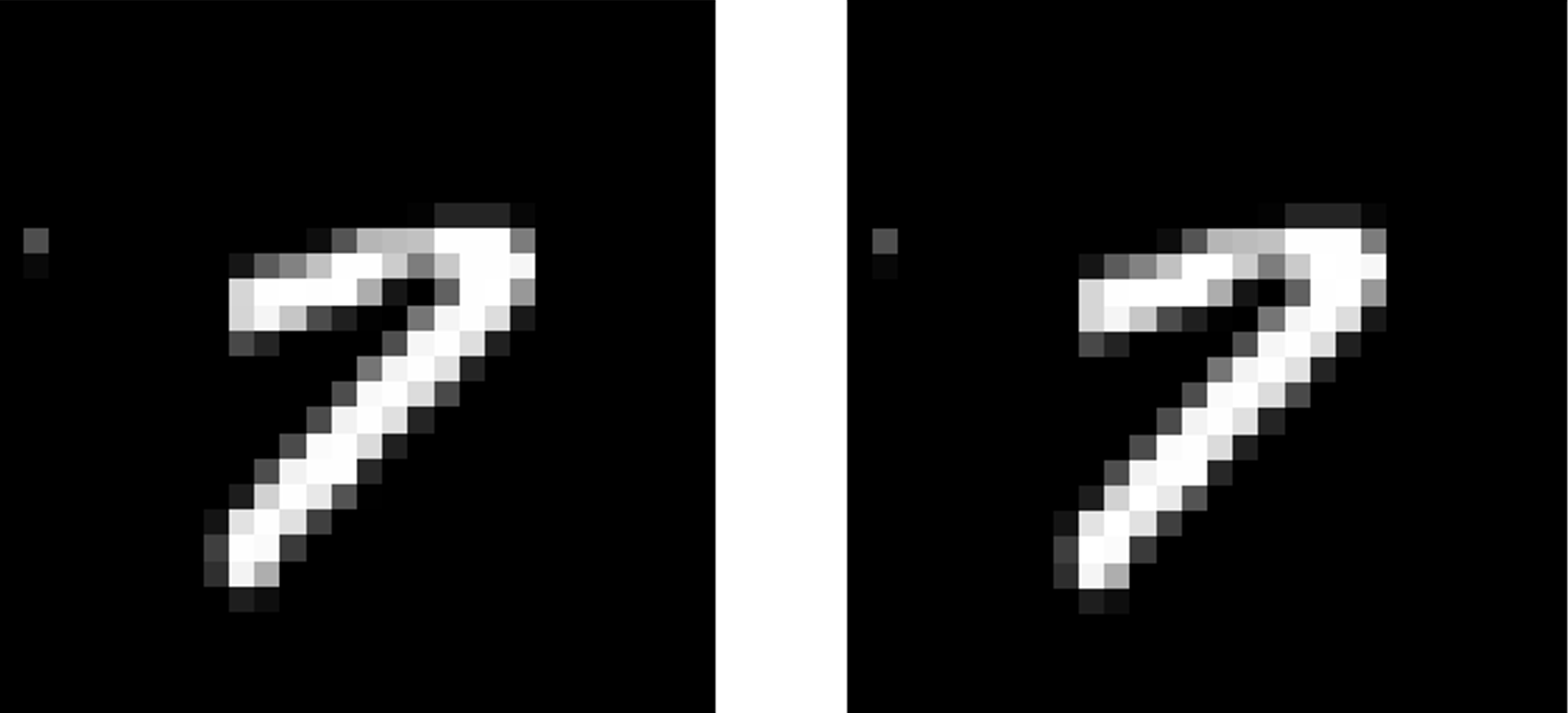}%
\caption{The original image (left) and a minimum adversarial example (right) found by \optimizer~($\epsilon=0.001$, adv.\ label: 1). \label{fig:example1}}
\end{minipage}
\hspace{6mm}
\begin{minipage}[t]{0.45\textwidth}
\includegraphics[width=5.5cm]{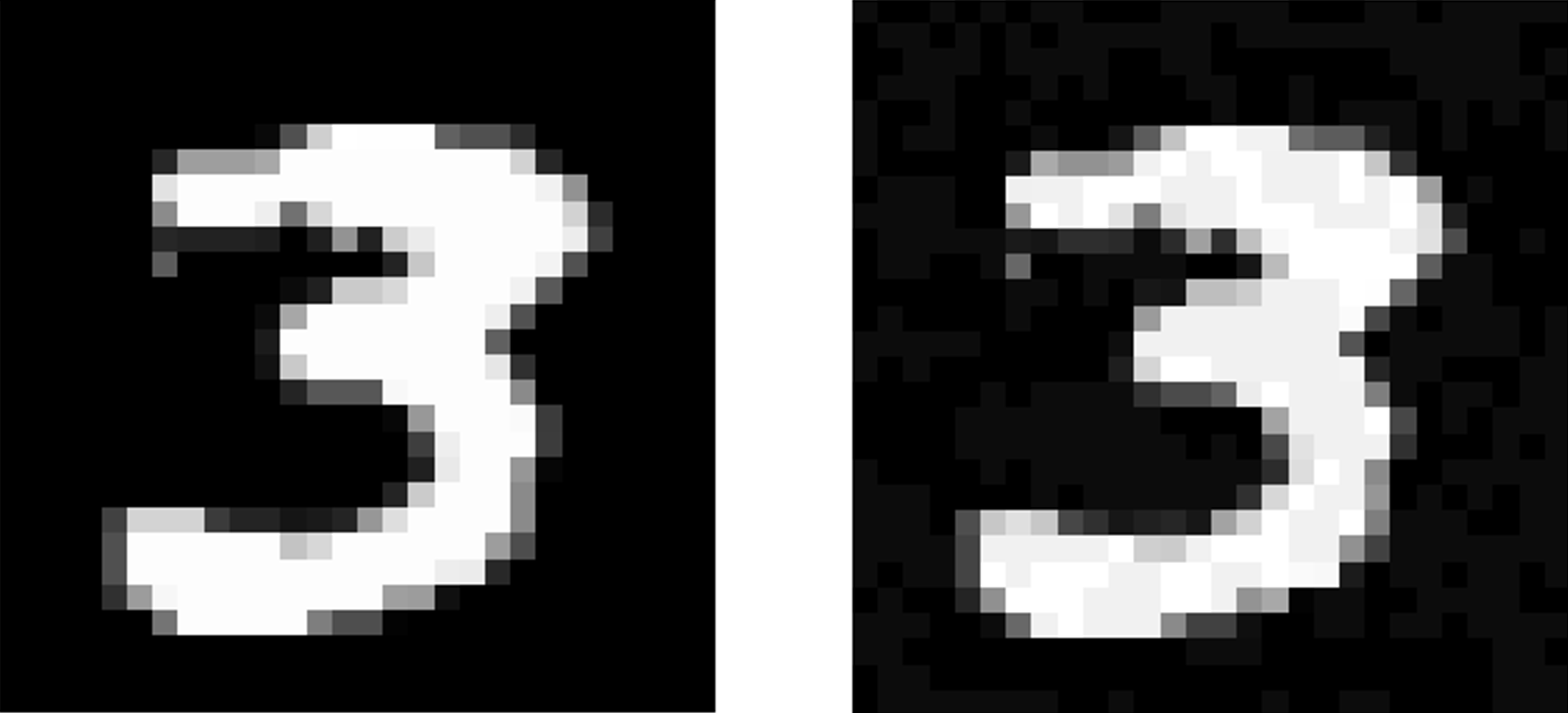}%
\caption{The original image (left) and a minimum adversarial example (right) found by \optimizer~($\epsilon=0.048$, adv.\ label: 7).\label{fig:example2}}
\end{minipage}

\end{figure}

In total, 522 out of the 785 benchmarks are solved. \optimizer~solved 12 infeasible queries, while MIPVerify solved 6. Figures \ref{fig:example1} and \ref{fig:example2} are two examples of the minimum adversarial inputs found by \optimizer~on MNIST1. In particular, a hardly discernible perturbation (0.001) can result in a misclassification in figure \ref{fig:example1}, while it requires a perturbation with $L_\infty$ norm of at least 0.048 (corresponding to 12 pixel-values) for the network to mistakenly classify the image as a ``7'' in figure \ref{fig:example2}, suggesting that the same network can exhibit very different local adversarial robustness properties in different input regions. 

\subsection{Objective Value Evaluation: Exact versus Approximate Optimization \label{subsec:approx-vs-exact}}

\begin{figure}
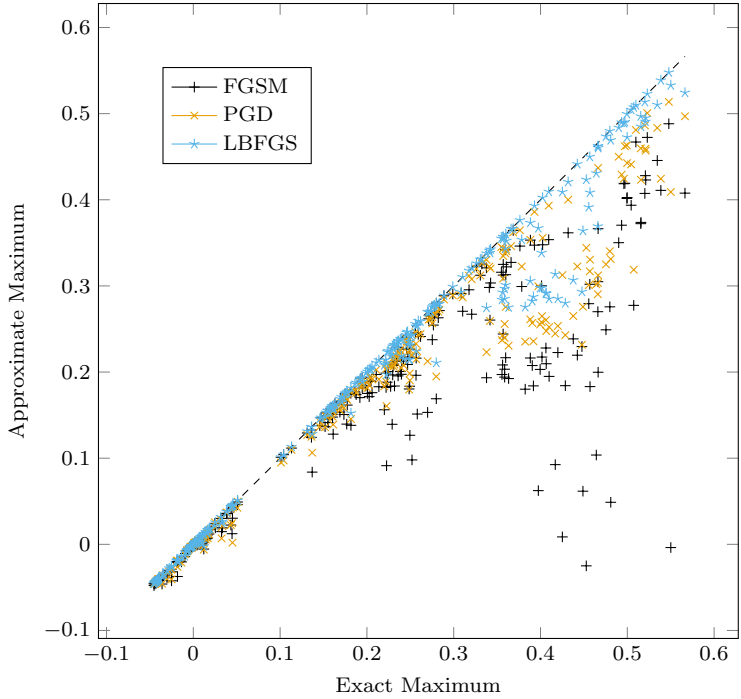


\begin{center}
\hspace{-1.5cm}


    \end{center}
    \caption{
    Exact versus approximate methods on output optimization problems. For each point, the position on the $x$ axis shows the true maximum found with MarabouOpt, MIPVerify, or MarabouBin while the position on the $y$ axis shows the objective value achieved by the approximate solvers (FGSM, PGD, or LBFGS). The dotted line is the line $y=x$. A perfect ``approximate'' optimizer that always achieves the true optimum would have points along the dashed line. As a result, the vertical gap between a point and the dotted line represents the gap between the exact maximum and the approximate value.
    }
    \label{fig:exactvsapproximate}
\end{figure}

In this subsection, we compare the performance of exact and approximate optimization approaches. We evaluate several tools on output optimization queries, and discuss how additional tools could be applied to minimum adversarial input queries.
\\\\
\xhdr{Output optimization queries} We run three approximate methods on all of the output optimization queries. We then compare the approximate values with the exact values and compare the runtimes. The exact values are obtained from MIPVerify, MarabouOpt, or MarabouBin, while the approximate values are obtained by the gradient-based optimizers, which includes LBFGS, PGD, and FGSM.
The values are compared in \cref{fig:exactvsapproximate}, with each point corresponding to a single query. The vertical distance between a point and the dotted line represents the gap between the approximate value and the true optimum. We observe that LBFGS most closely matches the exact values. We also observe that the gap between exact and approximate optimizers can be substantial, as evidenced by the points on the right side of the figure.
\Cref{tab:approximate runtime} provides statistics on the runtime of each approximate optimizer. Note that an issue with our implementation of LBFGS caused several queries to take much longer than the others, skewing its mean upwards. Comparing the medians, we see that FGSM was the fastest, followed by PGD, and then LBFGS. All three approximate solvers typically return within a second and should scale polynomially in the size of networks instead of the exponential scaling we observe for the exact optimizers.
These results suggest that for scenarios where a guarantee of optimality is not needed, approximate optimizers may often be ``close enough'' and provide a value in significantly less time. However, if strict guarantees are required, then an exact optimizer may be the better choice.

\xhdr{Minimum adversarial perturbation queries} A similar experiment could be performed for the minimum adversarial perturbation queries. PGD or LBFGS could be run with a loss that incorporates the size of the perturbation as in the work of \citet{yuan2019adversarial} to upper bound the minimum perturbation. An experiment like this was performed by \citet{carlini2017provably} to compare approximate and exact minimum adversarial perturbations. Alternatively, Fast-Lin, Fast-Lip, CROWN, and CNN-Cert are designed to provide a certified lower bound on the minimum perturbation which could be compared with the true minimum \citep{ weng2018towards,zhang2018efficient,boopathy2019cnn}. Although we did not run these experiments, they remain of interest for future work. 

\begin{table}[t!]
    \centering
    \caption{Statistics on the runtime of FGSM, PGD, and LBFGS for their solved output optimization query instances. \label{tab:approximate runtime}}
    \vspace{1ex}
    \begin{tabular}{@{}lllll@{}} 
        \toprule
        \textbf{Optimizer} & 
        \textbf{Minimum (s)} & \textbf{Mean (s)} & \textbf{Median (s)} & \textbf{Maximum (s)}\\
        \midrule
        FGSM & $0.1$ & $0.14$ & $0.13$ & $0.31$  \\
        PGD & $0.09$ & $0.23$ & $0.19$ & $0.96$ \\
        LBFGS & $0.02$ & $68.7$ & $0.35$ & $5893$\\
        \bottomrule
    \end{tabular}
\end{table}

\section{Related Work}
\label{sec:related work}
In this section, we summarize existing neural verification and neural optimization methods. We organize the discussion using the categorization of neural verifiers introduced by \citet{liu2019algorithms}. For optimization, we focus on works relevant to the two specific optimization problems addressed in this paper: output optimization problems and minimum adversarial input problems (see \cref{subsec:neural optimization problem}).
\\\\
\xhdr{Neural verification methods}
Neural verification methods check properties about the input-output relationship of a network. Specifically, given an input set and an output set, a verification method proves that all elements of the input set are mapped by the network into the output set. \Cref{sec:background} provides notation for this problem.

\citet{liu2019algorithms} separate verification algorithms into four categories: reachability, optimization, \searchandreach, and \searchandopt. In this work we make particular reference to reachability verifiers ExactReach and NNV \citep{xiang2017reachable, tran2020nnv}, optimization verifiers NSVerify and MIPVerify \citep{lomuscio2017approach, tjeng2017evaluating}, \searchandreach verifiers ReluVal and Neurify \citep{wang2018formal, wang2018efficient}, and \searchandopt verifiers Reluplex and Venus \citep{katz2017reluplex, katz2019marabou, botoeva2020efficient}. We refer the reader to \cref{subsec:approaches to verification} or the survey paper \citep{liu2019algorithms} for a more thorough description and discussion. Additionally, the 1st and 2nd International Verification of Neural Networks Competitions boast an extensive list of participating verification algorithms for a curious reader to explore \citep{VNN20,VNN21}. These tools typically fit into the categorization presented by \citet{liu2019algorithms}. Although we have not enumerated each solver here in favor of describing a few representative verifiers from each category, each verifier can be extended to perform optimization using the framework described in this paper.  \Citet{bunel2020branch} also provide a framework for neural verification in terms of a branch and bound search. Both frameworks are very useful for conceptualization of how different verification algorithms function and their common high-level structure.
\\\\
\xhdr{Exact optimization methods}
\label{subsec:exact optimization methods}
There are several existing techniques for solving general optimization problems on neural networks. These problems are directly encoded as mixed integer programs (MIP) \citep{wolsey1998integer, cheng2017maximum, fischetti2017deep, lomuscio2017approach} and solved by an MIP solver such as the open source solver GLPK \citep{makhorin2004glpk} or the highly optimized commercial solver Gurobi \citep{gurobi}. MIPVerify \citep{tjeng2017evaluating} employs a progressive bound-tightening preprocessing step and applies an MIP solver to the problem. This preprocessing pass has a significant impact on computation time, and a similar preprocessing pass is implemented in \optimizer.
Another approach treats existing neural verifiers as a black box, performing a bisection method using multiple verification calls. This method has been applied in adaptive stress testing of a control network \citep{julian2020validation} and to find minimally distorted adversarial examples \citep{carlini2017provably}. Optimization modulo theory solvers can also handle complex, non-convex optimization problems \citep{bjorner2015nuz, sebastiani2015pushing, sebastiani2020optimathsat}. However, these solvers are based on satisfiability modulo theories (SMT) technology and perform computation over real arithmetic, which has been reported to scale poorly compared to the less precise floating point optimizers \citep{katz2017reluplex}.
\\\\
\xhdr{Approximate optimization methods}
\label{subsec:approximate optimization methods}
There is also a rich body of literature on approximate techniques for both optimization problems, particularly within the field focused on adversarial example discovery. Generation of adversarial examples typically combines a local optimization method with heuristics and applies it to a neural network. Approximate approaches exchange optimality for computational efficiency and often focus on minimum perturbation problems \citep{szegedy2013intriguing, carlini2017towards}. \citet{yuan2019adversarial} offer a taxonomy of these techniques in their survey. We address the methods relevant for this work. LBFGS is a  quasi-newtonian optimization method that finds an approximate minimum perturbation adversarial input \citep{zhu1997algorithm, szegedy2013intriguing}. The verification tools Fast-Lin, Fast-Lip, CROWN, and CNN-Cert provide a lower bound on the minimum adversarial distortion and can be considered approximate optimizers as they yield an approximate value (lower bound) for the minimum distortion \citep{weng2018towards,zhang2018efficient,boopathy2019cnn}.
FGSM and PGD have been used to directly generate adversarial examples within a small input region, corresponding to our output optimization problem, instead of finding minimum perturbation adversarial examples \citep{goodfellow2014explaining, madry2017towards}. LBFGS can be used in this way as well.

\section{Conclusions}
\label{sec:conclusion}
This paper presents general strategies for extending different categories of neural verifiers to solve global optimization problems. It focuses on two classes of problems: output optimization and minimum adversarial perturbation problems. 
We extended the Marabou neural verifier to create an optimizer \optimizer and compared its runtime performance against the black-box bisection search and MIPVerify. 
We observed that on a significant majority of queries, \optimizer substantially outperformed a naive bisection based approach, showing the advantages of tight integration that can be achieved with proposed extension strategies. The comparison of
\optimizer and MIPVerify shows complementary performance, indicating that different optimizers have both strengths and weaknesses and that extensions of each verifier should be explored. 
Our comparison of the global optima found by these solvers to local optima from several adversarial attacks, while often similar, had some marked differences. 

Neural optimization problems have a wide variety of applications and merit deeper investigation. By encouraging the development of optimization techniques, we hope that more tools will be available to those working to verify safety critical systems.




Further work is needed to determine which verifiers will work best when extended to perform optimization tasks. This could consist of extending several more algorithms and comparing their performance against \optimizer~and MIPVerify. Additional work would also lie in incorporating some of the best practices from each technique into the others. For example, integrating MIPVerify's optimization-based preprocessing step \citep{tjeng2017evaluating} as we did with \optimizer~may provide other verifiers with a performance boost as well.

Incomplete solvers could also be used to obtain bounds on the optima for our optimization problems. For example, Ai2 and its updated version ERAN can compute an approximate reachable set for each layer \citep{GeMiDrTsChVe18,singh2018fast,singh2018boosting, balunovic2019certifying, singh2019beyond, singh2019abstract}. Maximizing an output objective over this approximate reachable set will give us an upper bound on the true optimum. For optimization-based methods, these bounds on the objective can be applied as constraints before solving. For some use cases, these bounds may provide enough information if an exact solution is not needed. 

There are several promising directions to improve \optimizer. Many of these involve reducing the depth of the tree the algorithm explores. Some directions include further developing our bound-tightening strategies, more intelligently choosing the node to split on, incorporating input splitting like in Marabou's \emph{Split-and-Conquer} strategy \citep{katz2019marabou}, and incorporating MIPVerify's preprocessing step. In addition, parallelization could be incorporated into the optimizer. Marabou has been able to use parallelism to great effect \citep{wu2020parallelization}. The same extensions could be explored with \optimizer. 


\section{Declarations}
\textbf{Funding:} Funding in direct support of this work: DARPA under contract FA8750-18-C-0099.
\\ \\
\textbf{Conflicts of interest/competing interests:} The authors have no conflicts of interest to declare that are relevant to the content of this article.
\\ \\
\textbf{Availability of data and material:} 
The networks and datasets used for testing can be found at \url{https://github.com/castrong/NeuralOptimization.jl}.
\\ \\
\textbf{Code availability:}
The benchmarking framework is available at \url{https://github.com/castrong/NeuralOptimization.jl}. The Marabou verifier can be found at \url{https://github.com/NeuralNetworkVerification/Marabou} with the optimization extension at \url{https://github.com/castrong/Marabou/tree/opt_branch_7_30}. A wrapper to run MIPVerify on these benchmarks is located at \url{https://github.com/castrong/MIPVerifyWrapper} and the original implementation is located at \url{https://github.com/vtjeng/MIPVerify.jl}. Implementations of adversarial attacks can be found at \url{https://github.com/jaypmorgan/Adversarial.jl}.

\section*{Conflict of interest}
The authors have no known conflicts of interest.

\begin{acknowledgements}
We would like to acknowledge support from Tomer Arnon, Christopher Lazarus, Changliu Liu, Ahmed Irfan, Chelsea Sidrane, Jayesh Gupta,  Alex Usvyatsov, Rianna Jitosho, Eric Luxenberg, and Katherine Strong.
\end{acknowledgements}

\bibliographystyle{spbasic}      
\bibliography{References}         


\end{document}